# Multitask Learning with Single Gradient Step Update for Task Balancing

**Sungjae Lee**

Department of Industrial and Systems Engineering, Dongguk University - Seoul

30 Pildong-Ro 1-Gil, Jung-Gu, Seoul 04620, South Korea

Tel: +82-2-2260-3840

sungjaelee@dgu.ac.kr

**Youngdoo Son**[*]

Department of Industrial and Systems Engineering, Dongguk University - Seoul

30 Pildong-Ro 1-Gil, Jung-Gu, Seoul 04620, South Korea

Tel: +82-2-2260-3840

youngdoo@dongguk.edu

\* Corresponding author

# Multitask Learning with Single Gradient Step Update for Task Balancing

**Abstract**

Multitask learning is a methodology to boost generalization performance and also reduce computational intensity and memory usage. However, learning multiple tasks simultaneously can be more difficult than learning a single task because it can cause imbalance among tasks. To address the imbalance problem, we propose an algorithm to balance between tasks at the gradient level by applying gradient-based meta-learning to multitask learning. The proposed method trains shared layers and task-specific layers separately so that the two layers with different roles in a multitask network can be fitted to their own purposes. In particular, the shared layer that contains informative knowledge shared among tasks is trained by employing single gradient step update and inner/outer loop training to mitigate the imbalance problem at the gradient level. We apply the proposed method to various multitask computer vision problems and achieve state-of-the-art performance.

**1 Introduction**

Convolutional neural networks (CNNs) have achieved remarkable success in various computer vision tasks such as image classification, object detection, semantic segmentation, and depth estimation [1-11]. Although these computer vision tasks were related to each other, most existing researchers have studied them individually. Recent researchers have shown that performing related tasks simultaneously is more efficient in memory usage and inference speed and can improve the performance of individual tasks [1-3].

The method of performing multiple tasks simultaneously is called multitask learning. Multitask learning is a methodology to improve generalization performance with informative knowledge shared between tasks. For successful multitask learning, the shared layers should represent informative knowledge across multiple tasks, and the task-specific layers should be well fitted to each task. In other words, both types of layers in multitask networks need to be learned to satisfy each purpose. However, training a multitask network is generally more difficult than single task learning owing to the need to learn multiple tasks simultaneously.

In general, the loss function of multitask learning is formulated as a weighted sum of each task loss, and the multitask network is trained with back propagation algorithm [12] employing the total loss, which is the weighted sum of task-specific losses. Because the multitask network is updated by the total loss, gradients that have passed through each task-specific layer may cause conflicts in the magnitude and direction of the gradients of each task in shared layers. Consequently, when a specific task gives dominant impacts to the shared layers, the multitask network can be biased to a specific task or not be well trained.

In the existing research addressing the task imbalance problem in multitask learning, the researchers have mainly studied in the following three directions: network architecture design, dynamic loss weighting, and multi-objective optimization. First, the network architecture design is a method for designing multitask networks employing task hierarchy, feature fusion, and attention mechanism. Specifically, in these studies, researchers designed hierarchical multitask networks to assign priorities among tasks, reflected different complexity between tasks using low-,

mid-, and high-level feature fusion, or performed pixel-to-pixel tasks using attention methods [3,5,7]. Most studies on network architecture design have focused on their own specific combinations of tasks to design multitask networks; therefore, it is not trivial to apply the studies to general multitask learning problems. Second, the dynamic loss weighting method is to search the optimal loss weights of each task every time step [1-3]. Dynamic loss weighting balances the multiple tasks by adjusting the loss weight to prevent any specific task from having a dominant impact on training the multitask network. It has the advantage of not requiring the tedious work of searching the optimal loss weights manually. However, in actual training of deep networks, gradients that have passed several task-specific layers may behave differently than expected even if the loss weights are multiplied to each task-specific loss. Third, multi-objective optimization is a method based on Pareto optimization [4,8]. This line of research assumes that training the multitask network with the weighted sum of task-specific losses is not adequate when tasks are competing with each other. Therefore, Pareto optimization is employed to search optimal gradient weights for updating the multitask network toward a common gradient direction with each task loss. As a result, Pareto optimization is similar to dynamic loss weighting in that it is necessary to search for the optimal weights for each task. However, because the process requires solving the multi-objective optimization problems to find the optimal weights, additional assumptions or constraints are required to apply the researches to large multitask networks, and this reduces the scalability to complex tasks.

In this study, we propose an algorithm to solve the imbalance problem of multitask learning employing inner/outer loop training procedure and a single gradient step update inspired by gradient-based meta-learning. The proposed method allows a multitask network to learn shared representations with informative knowledge across tasks that are not biased toward a specific task, and task-specific layers to be well fitted to each task with an ordinary gradient descent update. The brief explanation of the proposed method of this study is as follows. In the inner loop of training procedure, the shared layers are temporarily updated by a single gradient descent for each task while task-specific layers are fixed. Then, in the outer loop, the task-specific losses are obtained using the temporary shared layers updated for each task, and the original shared layer is then updated based on the sum of these task-specific losses. Next, the task-specific layer is updated using an ordinary gradient descent while the shared layer is fixed. Our proposed method repeats the training procedure above so that the shared and task-specific layers are alternately updated, and it can balance the gradients among tasks without additional loss or gradient weights. In addition, the algorithm explicitly distinguishes and updates shared layers and task-specific layers separately, which have two different roles in multitask networks. The shared layer is updated to satisfy across tasks through a single gradient step update, and each task-specific layer is updated independently so that it is well fitted to each task. Consequently, the proposed method can easily be applied to typical multitask learning problems and is scalable to large multitask networks because it does not require additional optimization process to search for loss or gradient weights.

The key contributions of this study are as follows. First, inspired by gradient-based meta-learning, which has commonalities with multitask learning, we developed the proposed method by appropriately applying a single gradient step update and inner/outer loop training to multitask learning. The proposed method alleviates the task imbalance problem, an ongoing issue of multitask learning, at the gradient level. Second, to the best of our knowledge, we are the first to combine multitask learning and gradient-based meta-learning. Third, we applied the proposed method to both simple and complex real-world computer vision tasks to verify the performance of the algorithm and achieve state-of-the-art performance.

This paper is organized as follows. In Section 2, we introduce the literature review related to multitask learning and meta-learning. In Section 3, we describe the proposed method for training multitask learning. Section 4 presents the experimental design and result with interpretation. In Section 5, we provide a conclusion with a brief summary and explain future works.

## 2. Related Work

### 2.1 Multitask Learning

Multitask learning is an algorithm that learn several tasks at the same time, which is not only efficient in memory usage and computational cost but also can enhance generalization performance by employing shared information between tasks. Multitask learning has traditionally been a subject in statistics studies but has recently became popular in machine learning [13]. Most recently, researchers have studied multitask learning in deep learning for computer vision tasks such as multiclass classification, semantic segmentation, depth estimation, normal estimation, and pose estimation based on CNNs [1-11]. In addition, deep multitask learning is actively studied because it has similarities with transfer learning, continual learning [3], curriculum learning [2], and, recently, few shot learning [14]. However, it is a major challenge for multitask learning to be balanced across tasks as multiple tasks are trained simultaneously. The following studies have been conducted to solve this challenge.

First, network architecture design entails modifying the structure of neural networks for multitask learning to balance or prioritize among multiple tasks. In [5], the researchers designed a hierarchical network architecture to prioritize multitask networks according to task difficulty, and [7] proposed a structure to fuse CNN features under the assumption that the required low-, mid-, and high-level features of the network are different according to the complexity of each task. [3] performed pixel-to-pixel prediction tasks by adding attention layers parallel to SegNet [15].

However, since these studies on neural network architecture design were focused on specific pre-intended multitask learning problems, there are limitations in applying these research findings to other multitask learning problems. For example, [7] involved concatenating low-, mid-, and high-level CNN features for feature fusion and delivering them to fully connected layers. Then, intermediate values were processed into each task-specific fully connected layer to perform face recognition, landmark points, and gender prediction. Therefore, extending this model to pixel-to-pixel tasks such as semantic segmentation is not trivial. In contrast, [3] proposed a neural network architecture that specialized in pixel-to-pixel tasks by combining attention layers with SegNet. In short, research on neural network architecture has a limitation in applying to different general multitask learning problems because researchers have focused on specific combinations of multiple tasks.

Second, loss balancing, dynamically adjusting the loss weights for each task, has been studied to achieve balanced learning in a multitask network. In previous studies, [2] dynamically balanced the weights for task-specific losses in the loss function of multitask learning using homoscedastic uncertainty, and [1] adjusted the loss weights by considering the amount of change in the loss and the gradient magnitude of the last shared layer every time step. Similarly, [3], a variant of [1], also proposed a loss-weighting method that used only the task-specific losses without the gradients of multitask networks. These loss balancing methods also update the multitask network with a backpropagation algorithm using a weighted sum of task-specific losses. Therefore, the gradients from each task-specific loss are added at the last layer of the shared layers after passing each task-specific layer, calculated by the chain rule. However, in large multitask networks for complex problems, if the scales of task-specific losses are made similar in the loss function by adjusting the weights, the magnitudes of the gradients added at the last shared layer can be different after passing through several task-specific layers.

Third, multi-objective optimization assumes that training a multitask network with a weighted sum of task-specific losses is inappropriate when tasks are competing with each other. Therefore, multitask learning takes place by employing Pareto optimality, which prevents other tasks from deteriorating due to the learning of a specific task, rather than imposing a trade-off problem between tasks. [8] distinguished task-specific layers from shared layers and trained task-specific layers first. Then, the optimization problem was solved to obtain the gradient weights for each task with a common descent direction and update the shared layer. [4] was a variant of [8] that allowed subtasks to be learned by assigning different preferences to tasks in multitask learning and then selected an appropriate network according to practitioner needs among the learned multiple multitask networks. Multi-objective optimization requires solving optimization problems to obtain additional weights, and as such, additional constraints or assumptions are needed to apply to large CNNs for complex real-world applications. Also, prioritizing a specific task increases exponentially as the number of tasks increases [4], so assigning task preferences in this study can be considered a grid search for loss weights.

## 2.2 Gradient-based meta-learning

Meta-learning extracts information across tasks from prior learning and enables more effective learning of novel tasks. One of the popular approaches of meta-learning is to learn the update rule of the task learner through the meta-learner [16-18]. Another famous approach is gradient-based meta-learning, which has been actively studied recently. This approach allows a single learner to quickly adapt to a novel task with only one or a few updates at the test time through gradient descent [14,19,20]. Gradient-based meta-learning is closely related to the method we propose in that both methods attempt to learn a network that well represents the information shared among the tasks.

Model-agnostic meta-learning (MAML) [14] is a breakthrough algorithm in gradient-based meta-learning that can be adapted to any learning models trained based on gradient descent. MAML learns sensitive initial parameters that enable quick adaptation to new tasks with only one or a few instances. Learning method in MAML consists of two phases, an inner and an outer loop. In the inner loop, the network performs task-specific fine tuning for a few instances of each task and then, in the outer loop, the network conducts the meta-update to minimize the sum of the loss of each task. MAML accumulates knowledge across tasks through previous experiences and enables the network to quickly adapt to a novel task without over-fitting through accumulated knowledge at test time. Recently, several variants of MAML have been proposed, and the following studies are also closely related to our work. [19] proposed an algorithm that reduces the number of parameters to be updated at the test time by determining the parameters to be learned by meta-update, unlike MAML, which updates the entire network for both the inner and outer loops. [20] proposed adding context parameters to the network and updating only these parameters at test phase. The variants of MAML improved predictive performance by introducing these task-specific parameters. Therefore, when multiple tasks exist, it is necessary to explicitly distinguish shared parameters across tasks and task-specific parameters for each task in the training procedure. Explicitly distinguishing the two parameters can actually enhance predictive performance as the subsets of parameters for each task can be learned for their own purpose.

Multitask learning learns a model that performs multiple tasks simultaneously by extracting shared knowledge across tasks. On the other hand, gradient-based meta-learning learns a model with a large number of tasks and can be adapted to a novel task by updating the whole network [14] or task-specific parameters [19,20], using a few instances at the test time. Although multitask learning and gradient-based meta-learning have different purposes of extracting and using shared knowledge from multiple tasks, they both extract shared information among the tasks. In addition, both methods have parameters that play different roles: shared parameters representing the knowledge across tasks, and task-specific parameters fitted only to specific tasks. However, to the best of our knowledge, although there are a few studies on combining meta-learning and multitask learning [21,22], there are none on combining gradient-based meta-learning and multitask learning, although they have the similarities above. Therefore, we propose an algorithm that solves the imbalance problem of multitask learning by properly combining these two methods.

## 3 Proposed method

In this section, we introduce our proposed method, which alleviates the imbalance problem of multitask learning. The proposed method appropriately applies the gradient-based learning method of MAML and its variants to multitask learning. In particular, single gradient step update and inner/outer loop training procedure can solve the problem that the shared layers are biased toward a specific task and enhance the predictive performance of the multitask network.

### 3.1 Notation and Problem Formulation

We consider a multitask network $f_\theta(\cdot)$ for all $T$ tasks with parameters of the shared layers $\theta_s$ and parameters of the task-specific layers $\theta_t^1, \theta_t^2, \ldots, \theta_t^T$ for each task. Input **x** passes through shared layers and task-specific layers and derives outputs $\hat{y}_1, \hat{y}_2, \ldots, \hat{y}_T$, respectively. The task-specific losses, $L_1, L_2, \ldots, L_T$, are calculated by task-specific loss functions such as classification and regression losses with the outputs and target data $y_1, y_2, \ldots, y_T$, and the total loss can be calculated as the weighted sum of the task-specific losses. The typical objective function of multitask learning is as follows:

$$\min_{\substack{\theta_s, \\ \theta_t^1, \theta_t^2, \dots, \theta_t^T}} \sum_i^T w_i L_i(f_{\theta_s, \theta_t^i})$$

, (1)

where $w_i$ is the loss weight for task $i$. The loss weights can be determined manually by practitioner searching [23,24] or automatically by loss balancing algorithms [1,2,3].

When training deep multitask learning through backpropagation algorithm [12] based on equation (1), gradients that have passed through each task-specific layer eventually meet in the shared layers. At this time, if the differences in the direction and magnitude of the gradients between tasks are large, the conflict of gradients can cause imbalance problems. In most of the existing studies, researchers attempted to alleviate the imbalance by manually searching or finding the optimal loss weights $w_i$ with loss balancing algorithms. However, $w_i$ may not work as expected when large multitask networks are employed for complex multitask learning problems, such as real-world computer vision tasks because $w_i$ is multiplied to each task-specific loss $L_i$ for the balanced learning across tasks at the last layer and the gradients should pass through several task-specific layers till they arrive at the last of the shared layers. Therefore, it is necessary to update shared layers, which are intended to have informative shared knowledge across tasks without overfitting to specific tasks, separately from updating a given task-specific layer. Thus, the proposed method trains shared and task-specific layers alternately, and it alleviates the imbalance problem between tasks without the additional process of searching for $w_i$.

**3.2 Multitask Learning via Gradient-based Meta Learning**

In multitask learning, the shared layers need to contain information that is shared across tasks without being biased towards a specific task, and the task-specific layers need to be well fitted to each task. The proposed method repeats the process of updating the two layers separately so that the shared layers and task-specific layers can match their roles. In particular, the shared layers are trained by employing the single gradient step update used in gradient-based meta-learning such as MAML and its variants [14,19,20] so that the shared layers are not over-fitted to a specific task and can learn informative shared representations. The shared layer is temporarily updated for each task in the inner loop by the single gradient step update of the gradient-based meta-learning algorithms and updated to satisfy across tasks in the outer loop. As multitask learning learns tasks of different difficulty with different loss functions, the training rates among tasks can be different. Therefore, it is necessary to make the training rates among tasks similar for balanced learning. For instance, if an easier task is trained relatively quickly, the proposed method should decrease the training rate; in contrast, if difficult tasks are trained relatively slowly, the proposed method should increase the training rates for these tasks. By adjusting the rates among tasks to be close, shared layers are prevented from receiving dominant impact on a specific task, so that the imbalance problem of multitask learning can be alleviated. To represent the informative shared layers, we propose an algorithm that solves the imbalance problem in multitask learning at the gradient level by considering a single gradient step update and inner/outer loop training procedure rather than searching loss or gradient weights in order to learn a complex nonlinear function deep CNN-based multitask network.

The detailed procedure of the algorithm is as follows. First, the loss for each task is measured, and the shared layer is temporarily updated with a single gradient step through gradient descent using each task loss while task-specific layers are fixed as in equation (2):

$$\theta_s^i \leftarrow \theta_s - \alpha \nabla_{\theta_s} L_i(f_{\theta_s, \theta_t^i})$$

, (2)

where α is a step size. This is equivalent to fine tuning at test time in gradient-based meta-learning algorithms. [25] showed that the single gradient step update can approximate deep network up to arbitrary accuracy. The loss of each task is calculated again using the temporarily updated shared layers for each task $\theta_s^i$ and the fixed task-specific layers $\theta_t^i$ to update the parameters in the shared layers with sum of the losses using equation (3):

$$\theta_s \leftarrow \theta_s - \beta \nabla_{\theta_s} \sum_i^T L_i(f_{\theta_s^i, \theta_t^i})$$

, (3)

where β is a step size. This is identical to the meta-update stage in the outer loop in gradient-based meta-learning algorithms. After the shared layers are updated, the task-specific layers are also updated by a single gradient step using equation (4):

$$\theta_t^i \leftarrow \theta_t^i - \alpha \nabla_{\theta_t^i} L_i\left(f_{\theta_s, \theta_t^i}\right)$$

, (4)

This training procedure is repeated every time step until the stopping criteria are satisfied.

The proposed method combines gradient-based meta-learning such as MAML and its variants with multitask learning. The differences from the existing gradient-based meta-learning algorithms are as follows. First, MAML performed gradient step updates for the entire network, but the proposed method only updates the shared layer. The concept of training by distinguishing the shared layer and the task-specific layer that play different roles in the proposed method is more similar to the variants of MAML [19,20]. Second, MAML and its variants are updated by resampling the mini-batch during fine tuning and meta-update, but the proposed method must perform updating to the same mini-batch because MAML and its variants focus on fast adaptation to novel tasks with only one or few instances; the proposed method, meanwhile, aims to conduct balanced training on given tasks in multitask learning. Third, the proposed method does not employ the second-order derivatives in equation (3) for reducing computational burden, unlike the gradient-based meta learning methods proposed in [14,19,20].

Fig. 1 illustrates the task-balancing effect of the proposed method for the situation of learning three tasks. Because multitask learning simultaneously learns tasks with different degrees of difficulty, the multitask network is trained through different gradient magnitudes between tasks as shown in Fig. 1(a). The black and red lines in Fig. 1 indicate the first gradients steps for temporary update of the shared layers, and the next gradient steps obtained using the temporarily updated shared layers for each task, respectively. The shared layers will be updated with different gradient magnitudes and directions from each task-specific loss function while each task is trained toward its own local minimum. In this situation, if the multitask network is immediately updated by the sum of task-specific losses without a single gradient step update as shown in Fig. 1(b), the shared layer can be biased toward the task with a relatively large gradient magnitude. This phenomenon can cause the shared layer to be over-fitted to a specific task, and this eventually results in the imbalance problem of multitask learning. In contrast, if the network is updated by the single gradient step-updated shared layers for each task as shown in Fig. 1(c), the training rate between tasks becomes closer. As shown in Fig. 1(a), if the task is trained on the area with positive curvature such as task 1, then it will be updated with the smaller gradient magnitude in the next step. On the other hand, if the task is trained on the area with negative curvature such as task 2, it will be updated with the larger gradient magnitude in the next step. Therefore, in the update of the shared layer after the single gradient step update as in equation (3), the training rates among tasks become closer at the gradient level compared with the direct update of the shared layers without temporary update using the single gradient step. By keeping the training rates among tasks close, the proposed method is more suitable for deep neural networks that have complex non-convex functions than the methods for explicitly assigning loss weights. Consequently, it can alleviate the problem of the shared layer being biased to a specific task. After all, the shared layers can be trained to well represent shared knowledge among tasks while not be over-fitted to a specific task, and the task-specific layers can be well fitted to each task.

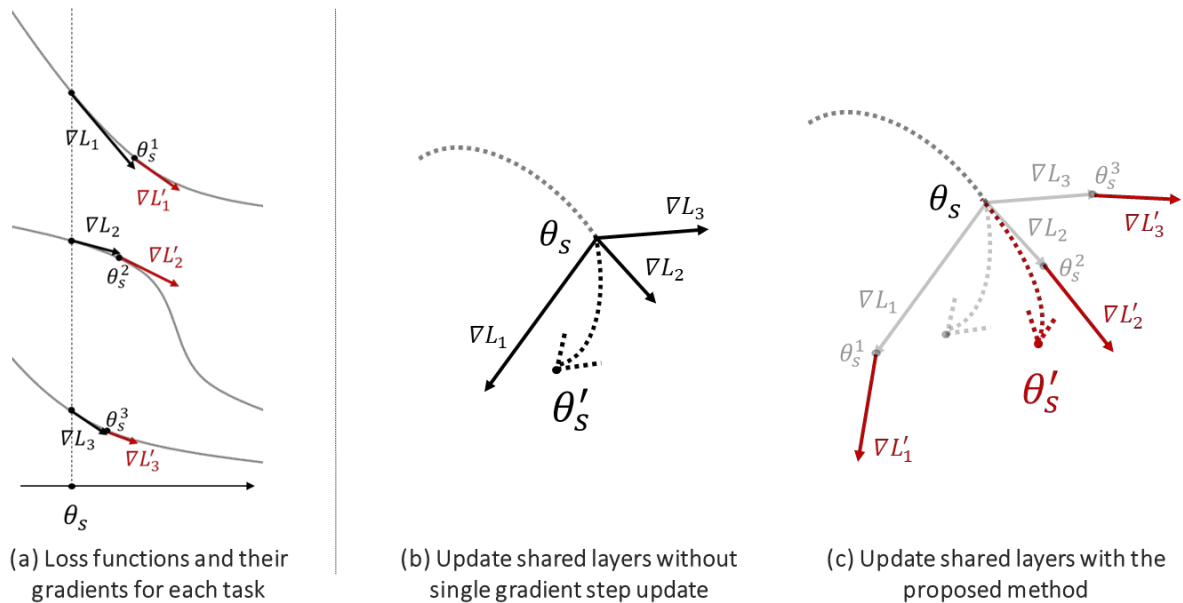

**Fig. 1.** Task-balancing effects of the single gradient step update in the proposed method.

The difference between the proposed method and existing multitask learning procedures can be expressed as the graphical models shown in Fig. 2. Fig, 2(a) displays learning a single task, where individual networks learn through task-specific loss function. Fig. 2(b) is ordinary multitask learning, which is trained by the sum of task-specific losses, and Fig. 2(c) adds the loss weights for employing the weighted sum of the task-specific losses. The optimal loss weights can be obtained manually or using grid search. Fig. 2(d) is the loss balancing method, the most frequently researched method on training multitask learning. The method searches the loss weights at every time step through an additional process and updates the multitask network using the weighted sum of the task-specific losses. Finally, Fig. 2(e) represents the update of shared layers in the proposed method. It temporarily updates a shared layer with a single gradient step for each task. Then, the parameters in the temporarily updated shared layers with the single gradient step, $\theta_s^1, \theta_s^2$, and $\theta_s^3$, derive each task-specific loss and update the shared layers. The proposed method can update the multitask network in the way we expect by updating the shared layer and task-specific layer separately, unlike with procedures of most existing studies. Therefore, after the shared layers are updated as shown in Fig. 2(e), the task-specific layers are updated with ordinary gradient descent while the parameters in the shared layers are fixed.

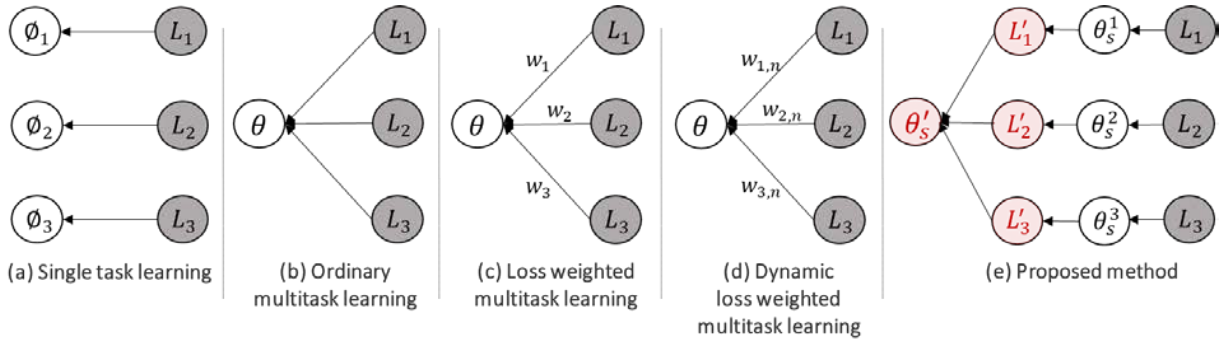

**Fig. 2.** Comparison of graphical models between existing studies and the proposed method (*n* denotes the time steps in (d)).

### 3.3 Algorithm and implementation

The procedure of the proposed method is summarized as follows. First, the shared layers are updated temporarily using a single gradient step for each task. Second, task-specific losses are calculated again with the temporarily updated shared layers, and then the shared layers are updated using these task-specific losses. Finally, the task-specific layers are also updated by a single gradient step using each task loss, and this learning procedure is repeated until the stopping criteria are satisfied. Algorithm 1 presents the detail of the proposed method:

---

**Algorithm 1:** Multitask learning with single gradient step update

---

1: randomly initialize $\theta = \{\theta_s, \theta_t^1, \theta_t^2, \dots, \theta_t^T\}$ and set learning rate $\alpha$, $\beta$

2: **for**

3:         sample mini-batch $D = \{x, y_1, y_2, \dots, y_T\}$

4:         **for all task $i$ do**

5:             compute $\nabla_{\theta_s} L_i(f_{\theta_s, \theta_t^i})$ using $x$, $y_i$, and $L_i$

6:             calculate single gradient step updated shared layer with gradient descent :
$\theta_s^i \leftarrow \theta_s - \alpha \nabla_{\theta_s} L_i(f_{\theta_s, \theta_t^i})$

7:         **end for**

8:         update $\theta_s \leftarrow \theta_s - \beta \nabla_{\theta_s} \sum_i^T L_i(f_{\theta_s^i, \theta_t^i})$ using $x$, $y_i$, and $L_i$

9:         **for all task $i$ do**

10:        update $\theta_t^i \leftarrow \theta_t^i - \alpha \nabla_{\theta_t^i} L_i(f_{\theta_s, \theta_t^i})$

11:        **end for**

12: **end for**

**Algorithm. 1.** Pseudo code of the proposed method

## 4 Experiments

We performed experiments on data sets of different multitask learning problems to compare the proposed method with existing studies. Most of the experimental setups are identical to the state-of-the-art algorithms for each dataset [3,4].

## 4.1 Multi-fashion-MNIST

The first experiments are on multi-MNIST, multi-fashion, and Multi-fashion and MNIST data sets, which are multitask versions of the MNIST and Fashion-MNIST data sets. The data sets are identical to those used in the state-of-the-art algorithm, Pareto MTL [4]. Multi-MNIST [26] was constructed by randomly picking two images from the original MNIST [27] data set and placing them in the top-left and bottom-right corners. Multi-fashion data were constructed in the same way from the original Fashion-MNIST [28]. Lastly, multi-fashion and MNIST were constructed by randomly picking a single image from the original MNIST and Fashion-MNIST. These multitask learning problems are composed of two classification tasks that classify 10 classes of numbers or clothes at the top-left and bottom-right corners.

We compare the proposed method with the existing methods such as ordinary multitask learning, Uncertainty [2], GradNorm [1], and Pareto MTL [4] for the data set. We also compare the proposed method with the split-only method, which updates the shared and task-specific layers alternatively without a single gradient step update. Uncertainty and GradNorm are algorithms that adjust loss weights dynamically, and Pareto MTL formulates multitask learning as a multi-objective optimization problem with task preferences. We do not use any loss weight adjustments for the ordinary and split-only methods or the proposed method. The model used in the experiment was the same structure as the multitask network based on LeNet [27] proposed in [4]. We use multiclass cross entropy as the loss function and stochastic gradient descent as the optimizer with learning rate 0.001. The data set was randomly divided into 120,000, 10,000, and 10,000 for the training, validation, and test sets, and we employed an early stopping algorithm using validation data for regularization.

Table 1 shows the experimental results for the multi-MNIST series data sets, and the best results for each task are boldfaced. The proposed method outperformed the other methods for all tasks except one task in the multi-fashion and MNIST data, in which Uncertainty was the best. The classification accuracies of the proposed method for the two tasks in multi-MNIST were 92.30 and 90.38, and those of multi-fashion were 82.41 and 81.78 for each task, which was the best performance. In multi fashion and MNIST, the accuracy of the proposed method for task 1, classifying the top-left object, was 93.56, which is lower than the Uncertainty task 1 accuracy of 94.14. However, the accuracy of Uncertainty for task 2 classifying the bottom-right image was 81.08, and that of the proposed method was 84.28, and this difference in accuracy for task 2 was larger than that in task 1. We interpret this finding as reflecting that Uncertainty is biased to task 1 due to imbalanced training. We also tested Pareto MTL by assigning extreme preferences to task 1 and task 2 as well as equal preference, and the algorithm with extreme preference vectors also showed significantly biased results for a specific task. In specific, in the case of extreme preference for task 2 in multi-fashion and multi-fashion and MNIST, Pareto MTL was biased to task 2 and showed accuracies of 81.89 and 85.03, respectively, while the accuracies for task 1 were 80.75 and 89.52. The accuracies for task 2 are higher than those for the proposed method, but task 1 accuracies are significantly lower than those for the other methods. In addition, the proposed method outperformed Pareto MTL even in the extreme preference cases for multi-MNIST.

**Table 1** Experimental results for multi-MNIST series.

| Type | Preference | Multi-MNIST | | Multi-fashion | | Multi fashion and MNIST | |
|---|---|---|---|---|---|---|---|
| | | Task 1 | Task 2 | Task 1 | Task 2 | Task 1 | Task 2 |
| Ordinary | | 91.63% | 89.67% | 81.97% | 81.25% | 93.52% | 83.84% |
| Uncertainty | | 91.79% | 89.08% | 81.78% | 81.21% | **94.14%** | 81.08% |
| GradNorm | | 89.65% | 86.79% | 80.82% | 77.88% | 85.34% | 81.27% |
| Pareto MTL | Equal | 91.76% | 89.49% | 82.12% | 81.35% | 92.68% | 84.24% |
| Split-only | | 91.52% | 89.52% | 81.88% | 81.36% | 93.01% | 83.67% |
| Proposed | | **92.30%** | **90.38%** | **82.41%** | **81.78%** | 93.56% | **84.28%** |
| | | | | | | | |
| Pareto MTL | Task 1 | 91.79% | 88.28% | 82.15% | 79.81% | 93.75% | 82.84% |
| Pareto MTL | Task 2 | 90.83% | 89.77% | 80.75% | **81.89%** | 89.52% | **85.03%** |

We also performed an additional experiment to check the effect of the single gradient step update method on the balanced learning. Multi-MNIST is composed of two tasks that employ the same loss function and classify the same objects, and therefore, the loss ratio for each task to total loss should be close to 0.5 if learning for the two tasks is balanced. However, in actual training, the loss ratio between tasks can vary. Fig. 3 shows the loss ratio of task 1 before and after a single gradient step is tracked at every time step. The black line denotes a loss ratio of task 1 loss $L_1(f_{\theta_s,\theta_t^1})$ in total loss $L_1(f_{\theta_s,\theta_t^1}) + L_2(f_{\theta_s,\theta_t^2})$ without the temporary update using a single gradient step, and the red line denotes that with a single gradient step update. We note that the red line approaches 0.5 more closely than the black line. Because the shared layers in the proposed method are updated by using a loss after a single gradient step update, corresponding to the red line, the model can be trained balanced between tasks.

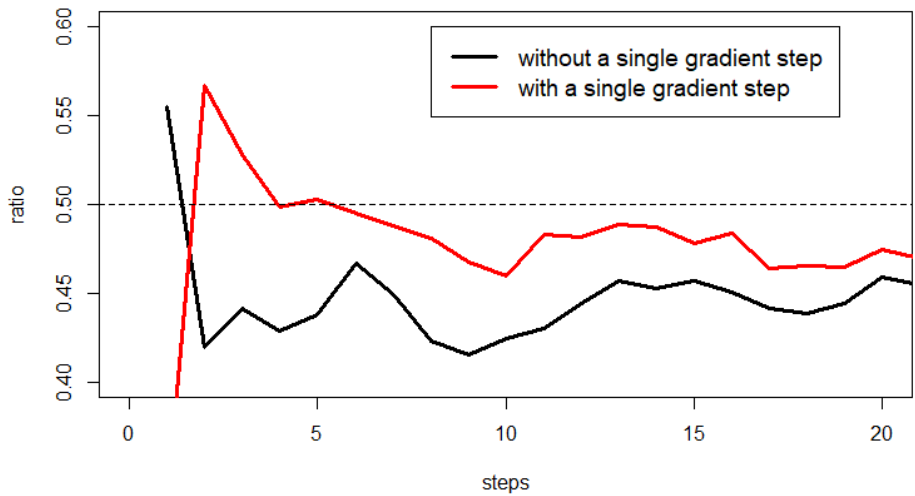

**Fig. 3.** Loss ratio of task 1 for multi-MNIST.

## 4.2 Real-world computer vision data sets

We conducted experiments on two complex real-world data sets that were identical to those used in the experiments of MTAN [3]. First, the CityScapes [29] data set, composed of high-resolution street-view images, contains tasks of semantic segmentation with seven classes and depth estimation. Second, NYUv2 [30] comprises RGB-D indoor scene images with the tasks of semantic segmentation, depth estimation, and surface normal prediction. For the semantic segmentation, depth estimation, and surface normal prediction tasks, we used pixel-wise cross entropy loss, $L_1$-loss, and element-wise dot products, respectively. For both data sets, we compared ordinary multitask learning, Uncertainty [2], GradNorm [1], and MTAN [3], the state-of-the-art algorithm for these data sets, with the proposed method, and we also compared the proposed method with Split-only model. The architectures of ordinary multitask learning, Uncertainty, GradNorm, Split-only, and the proposed method are based on SegNet [15]. We consider the encoder as the shared layers, and we construct the same number of decoders, the task-specific layers, as the number of tasks. Because MTAN is a specialized structure that combines attention layers parallel to SegNet, we used the same structure proposed in the original paper [3]. In the case of MTAN, we used the equal weighting method and used the results reported in the paper for comparison.

In the case of CityScapes, Split-only model and the proposed method showed the best performance on all measures as shown in Table 2. Split-only showed better results in the semantic segmentation task than the proposed method. We interpreted that it was sufficient to train by splitting the shared layer and task-specific layers, because semantic segmentation and depth estimation are mutually helpful for each other [31]. In addition, the state-of-the-art algorithm MTAN with equal weighting showed relatively poor results in depth estimation, and we interpreted that finding as the network's being biased toward the semantic segmentation task. Fig. 4 shows the qualitative analysis of CityScapes. The proposed method produced clearer and more accurate outputs than did ordinary multitask learning for the regions indicated by red boxes.

Table 2 The experimental results for CityScapes

| Type | Semantic segmentation | | Depth estimation | |
| --- | --- | --- | --- | --- |
| | Higher better | | Lower better | |
| | mIoU | Pixel accuracy | Absolute error | Relative error |
| Ordinary | 52.47 | 90.59 | 0.0147 | 24.62 |
| Uncertainty | 49.62 | 90.13 | 0.0146 | 24.22 |
| GradNorm | 46.62 | 87.71 | 0.0185 | 32.77 |
| MTAN | 53.04 | 91.11 | 0.0144 | 33.63 |
| Split-only | **54.09** | **91.18** | 0.0141 | 23.94 |
| Proposed | 53.80 | 90.96 | **0.0140** | **22.99** |

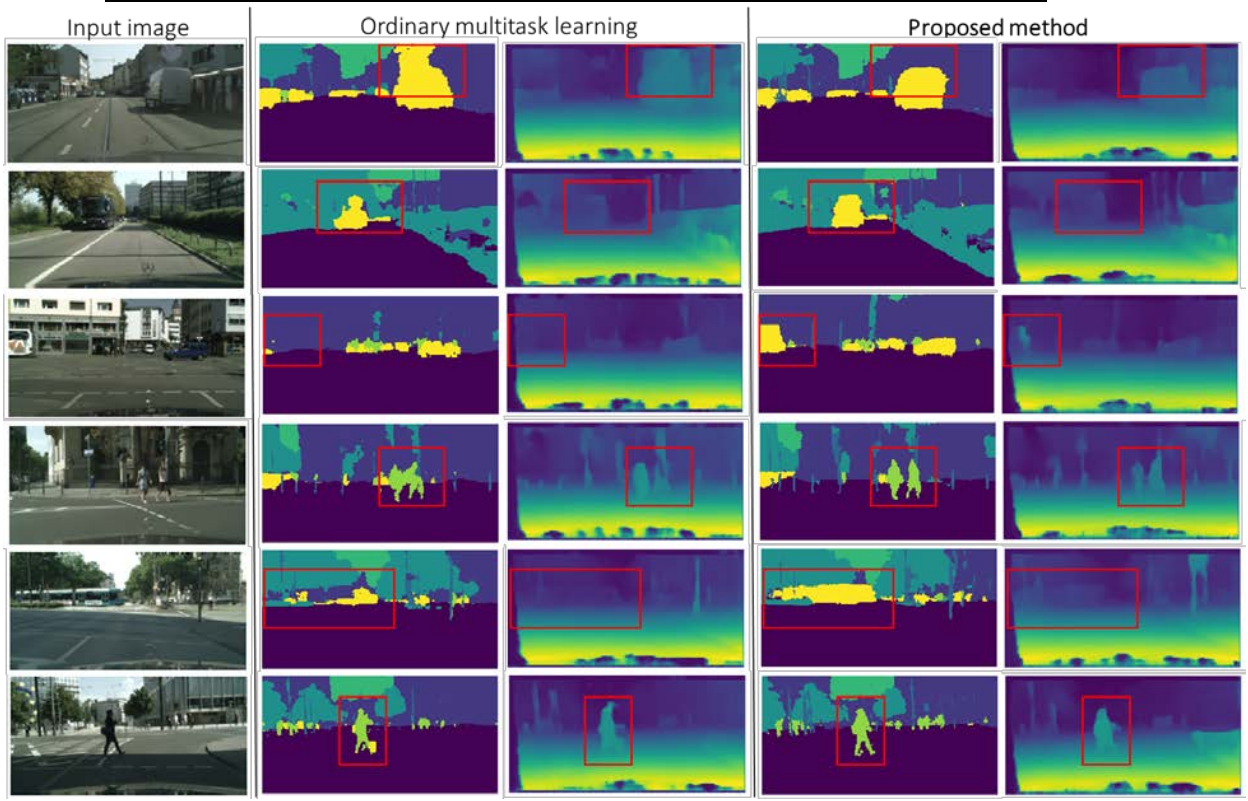

**Fig. 4.** Qualitative analysis, the actual outputs of ordinary multitask learning and proposed method for CityScapes data set.

In the case of NYUv2, the proposed method showed the best performance on 8 out of 9 measures as shown in Table 3; only for the absolute error of depth estimation, MTAN showed the best result. Interestingly, Split-only, which showed competitive performance in CityScapes, showed worse results than the proposed method for NYUv2. Our interpretation of these results is that it is not enough to alleviate the imbalance problem between tasks simply by training shared layer and task-specific layers separately because NYUv2 is composed of more complex data and surface normal prediction has been added as a task.

**Table 3** The experimental results for NYUv2

| Type | Semantic segmentation | | Depth estimation | | Surface normal | | | | |
|---|---|---|---|---|---|---|---|---|---|
| | Higher better | | Lower better | | Lower better | | Higher better (within $r°$) | | |
| | mIoU | Pixel accuracy | Absolute error | Relative error | Mean | Median | 11.25 | 22.5 | 30 |
| Ordinary | 17.44 | 54.92 | 0.6059 | 0.2573 | 31.39 | 25.36 | 23.42 | 45.61 | 57.22 |
| Uncertainty | 16.92 | 54.78 | 0.6099 | 0.2630 | 31.26 | 25.34 | 23.25 | 45.59 | 37.35 |
| GradNorm | 12.93 | 42.53 | 0.8048 | 0.3490 | 37.85 | 35.85 | 10.73 | 29.14 | 41.16 |
| MTAN | 17.72 | 55.32 | **0.5906** | 0.2577 | 31.44 | 25.37 | 23.17 | 45.65 | 57.48 |
| Split-only | 17.29 | 51.70 | 0.6478 | 0.2767 | 31.98 | 26.49 | 22.33 | 43.79 | 55.52 |
| Proposed | **19.49** | **55.95** | 0.6028 | **0.2514** | **30.83** | **25.12** | **23.70** | **45.90** | **57.72** |

## 5 Conclusion

In this study, we proposed a novel algorithm that solves the imbalance problem between tasks in multitask learning and that showed state-of-the-art performance in multiple data sets of multitask learning problems. The proposed method was inspired by the gradient-based meta-learning methods such as MAML and its variants because both multitask learning and gradient-based meta-learning extract the representation of shared information among the tasks rather than being biased toward certain tasks. In particular, the proposed method trains the shared layers separately from the task-specific layers, with single gradient step update and inner/outer loop training. To verify the effectiveness of the proposed method, we performed experiments on various simple and complex data sets, compared our proposed method with the existing methods including the state-of-the-art methods, and the results showed that the proposed method performed better than the other methods in most cases.

The proposed method can also be extended for additional studies as follows. First, in multitask learning, the parameters that are selected to be updated depending on the tasks may affect the predictive performance. Thus, it would be beneficial to find optimal subsets of the parameters in the shared layers to be updated based on the tasks for the proposed method, rather than updating all shared layers as in [32,33]. Second, research on an optimization method that can reduce the computation of the single gradient step updates in the inner loop as in [34] can be conducted to improve the efficiency of the proposed method. Finally, because the proposed method can be easily adapted to general multitask learning, it can be applied to various real-world multitask learning applications by being combined with other techniques [35,36].

**Acknowledgments**

This research was supported by the National Research Foundation of Korea (NRF) grant funded by the Korea government (MSIT: Ministry of Science and ICT) (No. 2020R1C1C1003425).